\documentclass[conference]{IEEEtran}
\input epsf
\usepackage{graphicx}

\begin{document}

\title{\LARGE Inter-Semantic Domain Adversarial in Histopathological Images}
\author{Nicolas Dumas, Louis-Oscar Morel, Sylvain Ladoire, Nathan Vinçon}

\author{\IEEEauthorblockN{Nicolas Dumas}
\IEEEauthorblockA{\textit{Ummon HealthTech}\\
Dijon, France \\
}
\and
\IEEEauthorblockN{Valentin Derangère}
\IEEEauthorblockA{\textit{Centre de Lutte contre le Cancer}\\
Dijon, France \\
}
\and
\IEEEauthorblockN{Laurent Arnould}
\IEEEauthorblockA{\textit{Centre de Lutte contre le Cancer}\\
Dijon, France \\
}
\and
\IEEEauthorblockN{\hspace{5em}}\\
\and
\IEEEauthorblockN{Sylvain Ladoire}
\IEEEauthorblockA{\textit{Centre de Lutte contre le Cancer}\\
Dijon, France \\
}
\and
\IEEEauthorblockN{Louis-Oscar Morel}
\IEEEauthorblockA{\textit{Ummon HealthTech}\\
Dijon, France \\
}
\and
\IEEEauthorblockN{Nathan Vinçon}
\IEEEauthorblockA{\textit{Ummon HealthTech}\\
Dijon, France \\
nathanvincon@ummonhealthtech.com}
}
\maketitle

\begin{abstract}
In computer vision, data shift has proven to be a major barrier for safe and robust deep learning applications. In medical applications, histopathological images are often associated with data shift and they are hardly available. It is important to understand to what extent a model can be made robust against data shift using all available data. Here, we first show that domain adversarial methods can be very deleterious if they are wrongly used. We then use domain adversarial methods to transfer data shift invariance from one dataset to another dataset with different semantics and show that domain adversarial methods are efficient inter-semantically with similar performance than intra-semantical domain adversarial methods.
\end{abstract}
\IEEEoverridecommandlockouts

\IEEEpeerreviewmaketitle

\section{Introduction}

\begin{figure*}
    \centering
    \includegraphics[scale = 0.5]{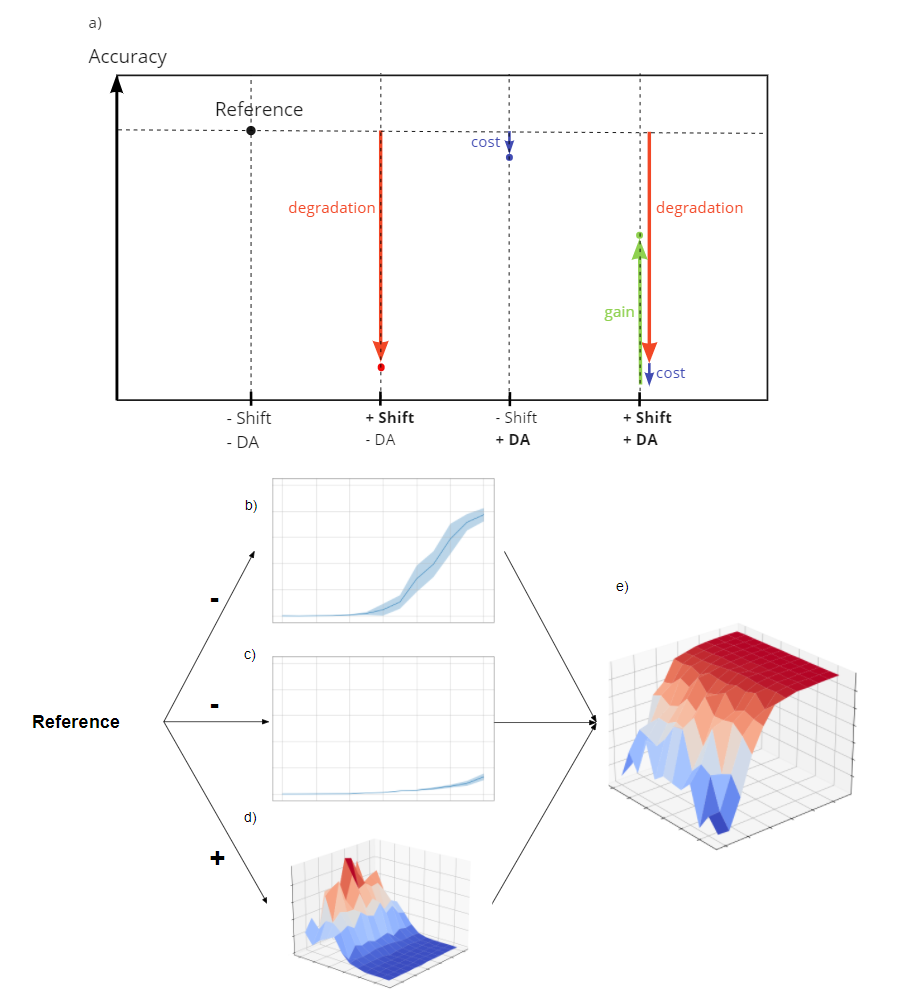}
    \caption{a) Diagram of the 4 pipeline configurations for calculating the reference, degradation, cost and gain. In the configuration with test shift and DA shift, the gain is obtained by the equation 
$$gain = accuracy - (reference - (degradation + cost))$$
b) Example plot of the degradation according to the test shift intensity 
c) Example plot of the cost according to the DA shift intensity
d) Example plot of the gain according to both test shift intensity and DA shift intensity. The x-axis is the test shift intensity, the y-axis is the score, the z-axis is the DA shift intensity
e) Example plot of the accuracy according to both test shift intensity and DA shift intensity. The x-axis is the test shift intensity, the y-axis is the score, the z-axis is the DA shift intensity. 
}
    \label{fig:model}
\end{figure*}

In computer vision, data shift (i.e. a shift in the data distribution) has proven to be a major barrier for safe and robust deep learning real-world applications, such as computer vision for autonomous vehicles \cite{filos_can_2020} \cite{alberti_idda_2020}, pose estimation \cite{zhang_keypoint-graph-driven_2021}, medical image segmentation and classification \cite{zhang_cross-denoising_2020} \cite{pooch_can_2019}. 
\\
\\
Among medical images, histopathological images are tissue sections stained and analyzed with a microscope by a pathologist to highlight features from the tissue related with diseases. These images are the gold standard for cancer diagnosis and still have a huge potential to improve clinical practices \cite{pronier_abstract_2020}. Some data shifts for histopathological images are known such as differences in the acquisition device parameters, differences in the staining and the multitude of parameters in the different steps of the histopathology slide preparation. However, risk may also come from a type of data shift that is not yet known.
\\
\\
Domain Adversarial (DA) \cite{ganin_domain-adversarial_2016} training has proven to be effective against data shifts notably in histopathological images \cite{lafarge_domain-adversarial_2017}. However, classic DA training configuration requires a sample from the targeted source of data (e.g. data from a new laboratory). This is generally not a problem as it only requires to pick samples from the new environment and train with the DA without requiring to label these new data. However, in clinical application, it is generally not possible to fine-tune a posteriori, as it would require clinically validating the model again. But clinical applications need a proven robustness to satisfy regulatory requirements.
\\
\\
In light of this DA technique and the robustness requirement of medical applications, an important question is whether we can use data with different semantics (e.g. flowers images and vehicle images have different semantics, prostate cancer and lung cancer images have close but also different semantics) as DA data. For example, can we use a multi-source dataset of prostate cancer for DA training while running a lung cancer classification task. In order words, we question the transferability of the domain adversarial process across tasks and image semantics. Transferability of DA training would imply that any task could gain generalization from a large dataset not only through classical transfer learning but also through DA transfer.
\\
\\
In this paper, we first investigate to what extent DA methods are beneficial and whether it can be deleterious. We then investigate to what extent DA methods can be transferred across datasets of different semantics (inter-semantic domain adversarial). Our contribution is :
\begin{itemize}
    \item We analyze the DA efficiency by describing 3 effects (Figure \ref{fig:model}) : the \textbf{cost} (i.e. negative difference of accuracy with baseline due to DA training), the \textbf{degradation} (i.e. negative difference of accuracy with baseline due to data shift), and the \textbf{gain} (i.e. positive difference with baseline after estimation of cost and degradation due to the consistency between data shift and DA). We further combine these effects in a regression model (see Overview of our approach) and show that a misuse of artificial domain shift such as color shift can be deleterious.
    \item We test to what extent DA training can be effective when the main task datasets and domain adversarial datasets are of different semantics. We show that DA can be transferred inter-semantically and that a small intensity of shift is sufficient to prevent most of the performance degradation due to the data shift.
\end{itemize}

\section{Background and Related Work}
In this Section, we review methods that were described to increase robustness and generalization over data shift.
\\
\\
Among these methods, we find :
\begin{itemize}
    \item Image augmentation \cite{faryna_tailoring_nodate} \cite{tellez_quantifying_2019} : the technique is easy to implement but can hurt performances if augmentation method is inadapted. Image augmentation also does not increase training time (or few), and does not increase prediction time.
    \item Stain and brightness normalization \cite{macenko_method_2009} \cite{khan_nonlinear_2014} \cite{vahadane_structure-preserving_2016} : the technique requires to infer a source and a target stain, then to transform the input image. It is easy to implement but can hurt performance if the stain inference step for source image is not robust enough, Ren at al. provided a solution using ensembles \cite{ren_unsupervised_2019}. Stain normalization can provide precise information about the stainings that can be later used for quality controls.
    \item GANs such as Stain-to-Stain Translation \cite{salehi_pix2pix-based_2020}, a pix2pix-based \cite{isola_image--image_2017} method or Cycle-GAN \cite{zhu_unpaired_2017}: they are efficient but complex, unstable and expensive techniques that increase prediction time.
    \item Domain Adversarial : a domain adaptation method where a DA branch is added after a feature extractor using a Gradient Reversal Layer, which prevents the top layer of the feature extractor from containing information about the domain considered irrelevant for the main task. DA can be seen as a disentanglement method of data shift over informative features. The DA training is directly targeting the model, therefore prediction time is not modified. Visual domain adaptation methods are reviewed in \cite{wang_deep_2018}.
\end{itemize}

\section{Material and Methods}
\subsection{Datasets}
We used MNIST and Fashion-MNIST datasets, both composed of 60,000 images with dimensions 28x28 (Figure \ref{fig:MNIST_dataset}).
\\
\\
We used 2 datasets of histopathological images. The first is CAMELYON \cite{litjens_1399_2018}, composed of 327,680 color images with dimensions 96x96x3 extracted from histopathological analyses of lymph node sections. Each image is annotated with a binary label indicating the presence of metastatic tissue. The second dataset, which we call TissueNet, is the dataset from the TissueNet challenge \cite{drivendata_tissuenet_nodate}. This dataset is composed of more than 5,000 images of uterine cervical tissue from 18 medical centers across France. The images are labelled on 4 levels depending on the grade of the cancer as follows :
\begin{itemize}
    \item 0 : benign
    \item 1 : low malignant potential
    \item 2 : high malignant potential
    \item 3 : invasive cancer \\
\end{itemize}

We used the RandomResizedCrop transformation of the torchvision python library to generate 60,000 images of dimensions 96x96x3.

\subsection{Domain Adversarial}
\begin{figure*}
    \centering
    \includegraphics[scale = 0.3]{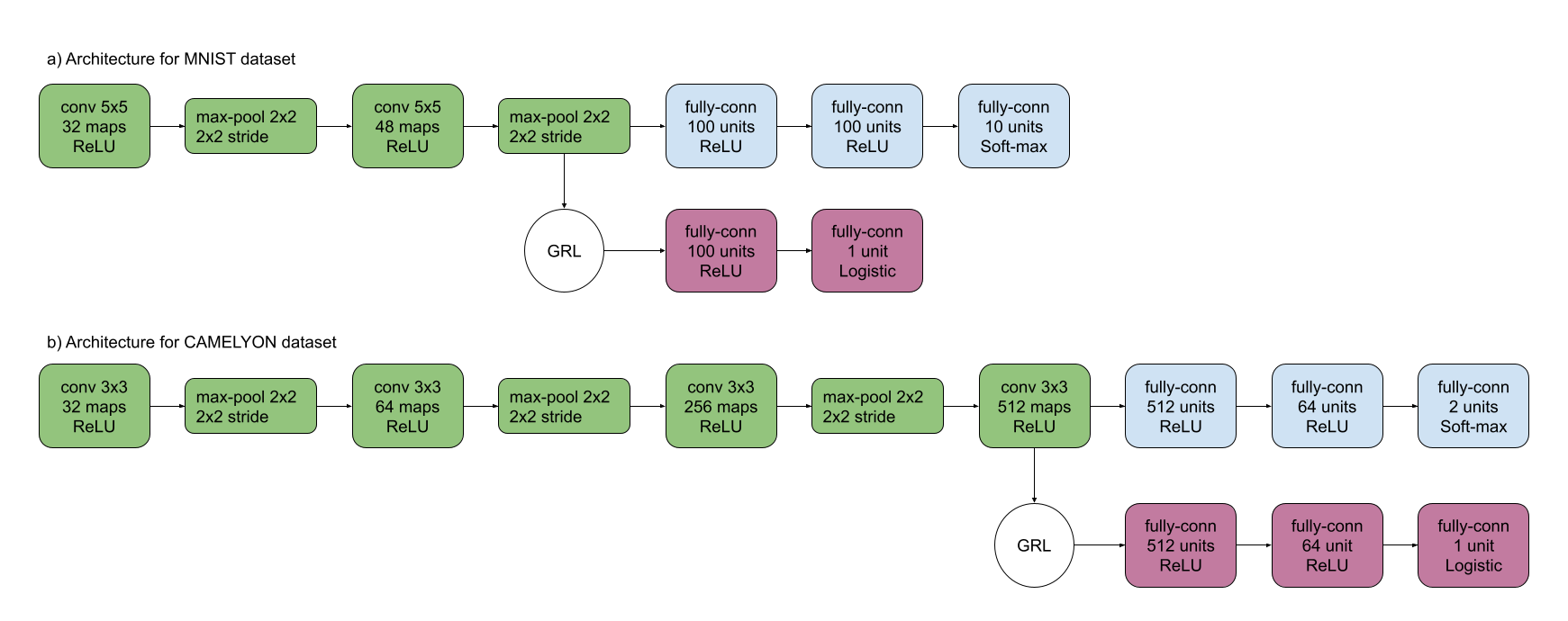}
    \caption{Architecture of the CNN used in the experiments. Boxes correspond to the layers. Green boxes correspond to the feature extractor, blue boxes to the label predictor and red boxes to the domain classifier. GRL is the gradient reversal layer.}
    \label{fig:architecture}
\end{figure*}

The proposed architecture is based on the one described in Ganin \textit{et al.} It includes a deep feature extractor and a deep label predictor, which together constitute a standard feed-forward architecture like a CNN. Domain adaptation is achieved by adding a domain classifier connected to the feature extractor via a gradient reversal layer (Figure \ref{fig:architecture}). By adding a domain classifier after the feature extractor architecture, we build the domain adversarial neural network (DANN). The domain classifier is trained with a mix of datasets from different domains (i.e. with different data shifts) labeled with their domain (Figure \ref{fig:training}).

\subsection{Data shift}
We tested both destructive shift (e.g. blur or noise), and domain shift (e.g. color shift and luminosity shift). The former aim to provide robustness against degraded input that can drastically hurt performance in a scenario where the model was trained on a curated dataset with high-quality images. The latter aim to provide invariance, therefore robustness, against input of the same quality but with natural data shift.
\\
\subsubsection{Noise}
The noise fonction is denoted by $N_i$ for $i$ over $[0, 12]$. Let $Im$ be an image represented by a float matrix with values between 0 and 1.  $N_i$ is defined by :
$$N_i(Im) = clip(Im + i*R, 0, 1)$$
with $R$ a random matrix of the same dimension as $Im$ and following a uniform distribution on $[0, 1]$
\\
\subsubsection{Blur}
Blur is done by convolving an image with a normalized uniform filter. The kernel is obtained by the following equation :
$$K = 1/(k*k) * M$$
with $M$ a matrix of 1 of dimension $k$ by $k$.
\\
\subsubsection{Color shift}
\begin{figure}
    \centering
    \includegraphics[scale = 0.12]{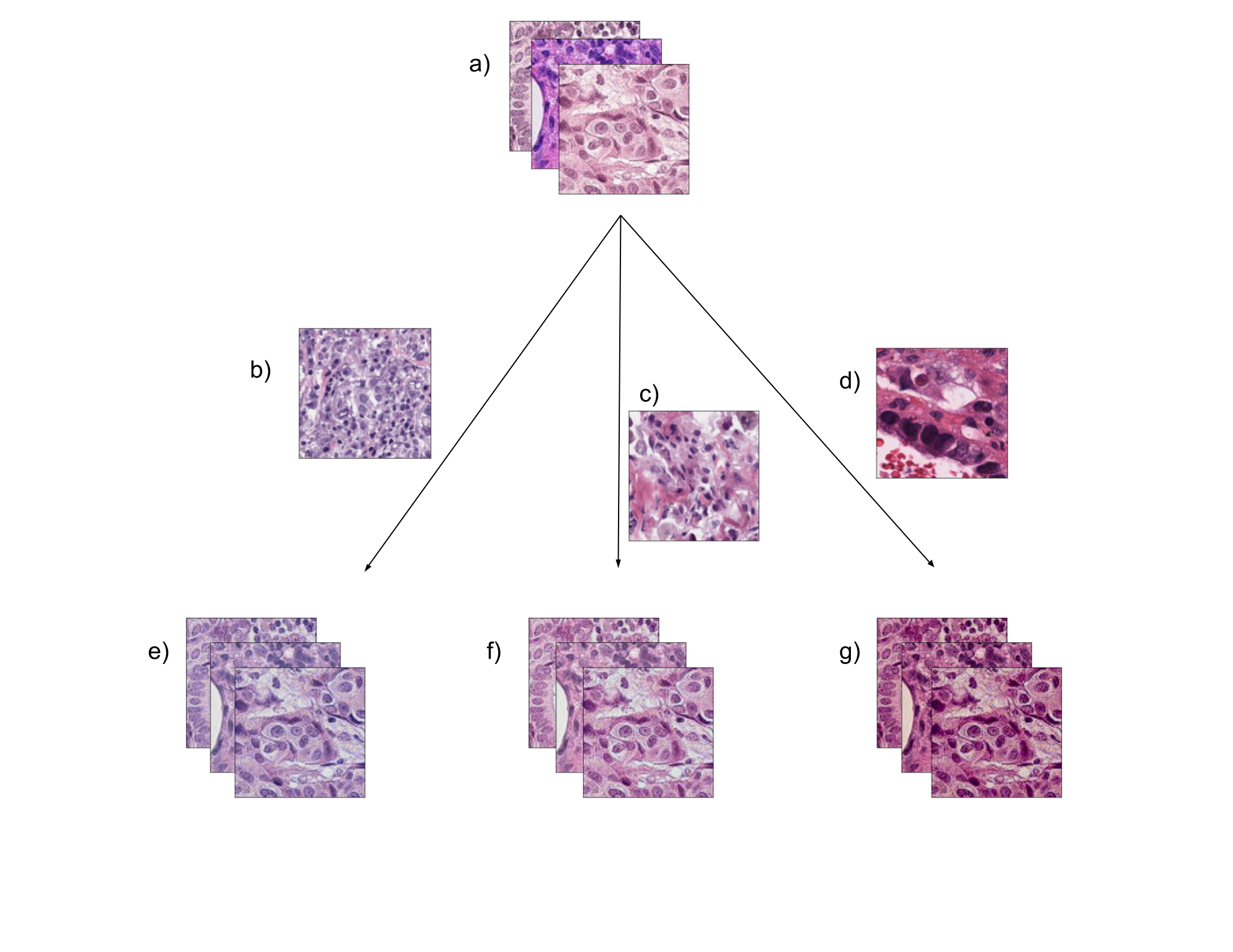}
    \caption{Image of the CAMELYON dataset transformed by stain normalization in different domains. a) original dataset.  e) f) and g) are the images of the CAMELYON dataset respectively normalized with b) c) and d) as target images.}
    \label{fig:dataset1}
\end{figure}

The color shift is based on the stain normalization method of reinhard \cite{reinhard_color_2001}. We convert the image in the LAB format, then we fit the mean and the standard deviation of the 3 channels. 
$$F(Im) =  (Im - avg(Im)) * (std(T) / std(Im)) + avg(T)$$
$Im$ is the original image, $T$ the target image , $avg$ is the average function and $std$ is the standard deviation. .
\\
More precisely, we use this normalization process as a color shift according to a reference image (Figure \ref{fig:dataset1}).

\subsection{Training}
The default training configuration requires 4 datasets (only 2 if we don’t run DA) :
\begin{itemize}
    \item the train and test datasets as for main classification task
    \item the two datasets used to train the DA (generally there is a data shift between these two DA datasets) \\
\end{itemize}

\begin{figure}
    \centering
    \includegraphics[scale = 0.25]{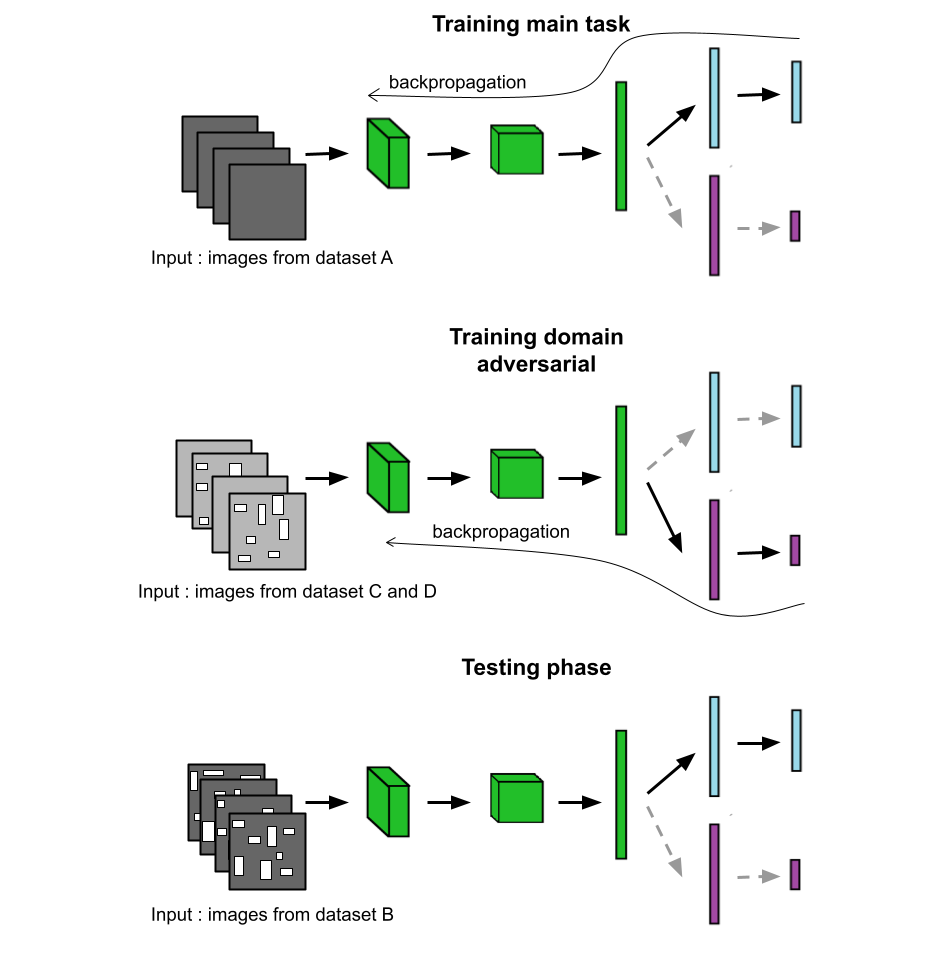}
    \caption{Diagram of the training and testing of the model. Green boxes correspond to the feature extractor, blue boxes to the label predictor and red boxes to the domain classifier. The black arrows show the path taken by the data. The main task training and domain adversarial training are made during the same phase. Dataset A is the training dataset, dataset B is the testing dataset, datasets C and D are the DA training datasets mixed for the training.}
    \label{fig:training}
\end{figure}

Note that the semantic of datasets for the main classification task and for the DA learning can be different. The training of the classification task and training of the DA are simultaneous (Figure \ref{fig:training}). Test shift refers to the data shift in the test dataset, and DA shift refers to the data shift between the two DA datasets.

\subsection{Regression model}
We used a regression model such that the data shift and DA effects are defined by 4 terms (Figure \ref{fig:training}):
$$Accuracy = reference - degradation - cost + gain$$
\begin{itemize}
    \item The reference is the performance of a raw CNN tested without data shift in the test set.
    \item The degradation term corresponds to the difference of accuracy with the reference due to a data shift in the test dataset. 
    \item The cost term corresponds to the difference of accuracy with the reference due to the DA training without any degradation. 
    \item The gain term corresponds to the difference of accuracy with a model after applying degradation and cost to the reference, such that the previous equation is satisfied.
\end{itemize}

\section{Experiments and Analyses}
\subsection{Characterization of the effect of domain adversarial training over model performance against noise perturbation}
\begin{figure}
    \centering
    \includegraphics[scale = 0.3]{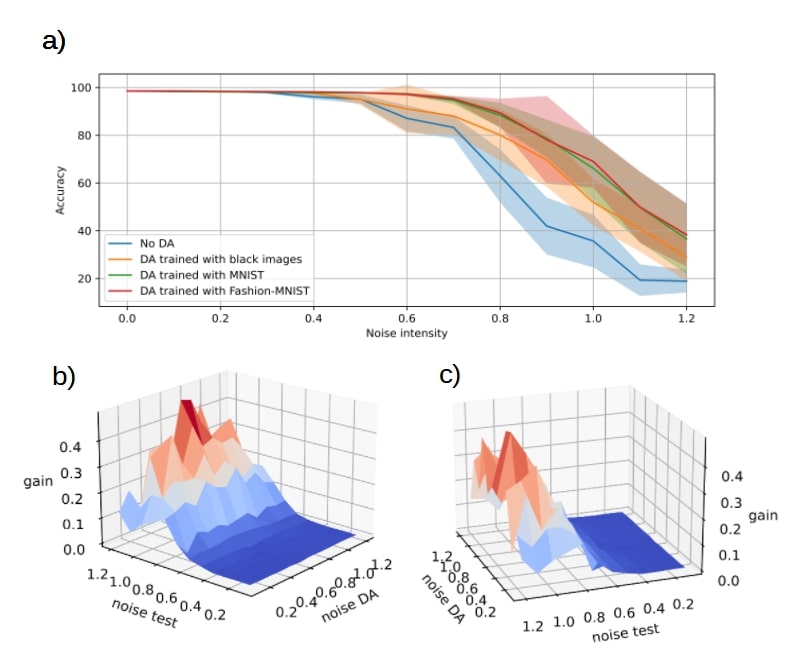}
    \caption{a) Plot of the accuracy for image classification of MNIST dataset as a function of noise intensity. For each curve, DA is trained with a different dataset. The noise intensity in DA is the same as in the test dataset. b) Representation of the gain with DA trained with MNIST dataset and noise as data shift. c) Same image as b) with different orientation.}
    \label{fig:MNIST}
\end{figure}

We first characterized the influence of data shift and DA training over the model performance. We used a regression model to link model performance to the data shift and DA training. This model is composed of 4 terms, the reference, the degradation due to data shift, the cost due to DA and the gain due to DA after degradation and cost (see Materials and Methods).
\\
\\
Using MNIST dataset and noise as data shift, the reference model reached 0.98 accuracy. We found a sigmoidal relation between noise intensity and performance degradation and the accuracy falls to 0.20 (degradation of 0.78) with a noise intensity of 1.2. We found that DA training had no deleterious impact over the model accuracy, therefore DA has no cost. We model the bivariate DA gain function as the product of two univariate functions: a gaussian function dependent on the noise intensity, and a quadratic function dependent on the DA intensity (Figure \ref{fig:MNIST}).
\\
\\
Interestingly, gain depends heavily on the data shift intensity but is almost invariant to DA intensity, suggesting that most of the DA benefits can be reached at very low DA intensity, this is further discussed in the Discussion. The maximum gain of 0.50 accuracy is reached with an intermediate noise intensity (1.1 ) and a low DA intensity (0.5).

\subsection{Inter-semantic transferability of the domain adversarial training in the MNIST dataset against noise perturbation}

\begin{figure}
    \centering
    \includegraphics[scale = 0.3]{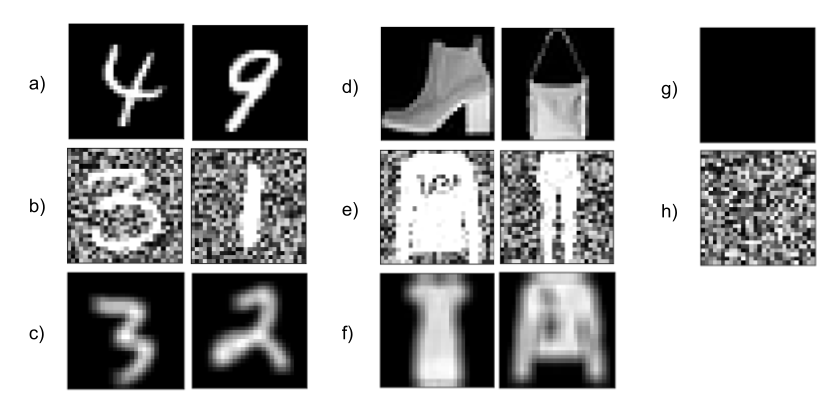}
    \caption{Images extracted from the different datasets used a) MNIST b) Noised MNIST c) Blurred MNIST d) Fashion-MNIST e) Noised Fashion-MNIST f) Blurred Fashion-MNIST g) Black image h) Noised black image}
    \label{fig:MNIST_dataset}
\end{figure}

Similar tests were performed using Fashion-MNIST instead of MNIST for the DA training, noise has been applied as previously. In this configuration, the DA training runs inter-semantically.
\\
\\
Training using DA training on Fashion-MNIST shows results that are close to the previous experiment (Figure \ref{fig:MNIST_blur}). We find a reference accuracy of 0.98 and a sigmoidal relation between noise intensity and performance degradation, and a maximum degradation of 0.78 with a maximum noise of 1.2. We find no cost associated with DA training. The gain is also the same as the configuration with MNIST in the DA datasets, modeled by the product of a gaussian function and a quadratic function, and mainly sensitive to noise intensity. The maximum gain of 0.45 accuracy is reached with an intermediate noise intensity as test shift (0.9) and an intermediate noise intensity as DA shift (0.9).
\\
\\
This experiment shows that using DA inter-semantically shows similar DA gain compared to intra-semantical DA. It suggests that the DA training for noise is entirely transferable between MNIST and Fashion-MNIST, such that Fashion-MNIST can be used equivalently to MNIST for the DA training in a MNIST classification task.
\\
\\
We performed similar tests using black images (Figure \ref{fig:MNIST_dataset}) instead of the Fashion-MNIST dataset as DA datasets and showed it has significant DA gain of 0.27. However, the gain is smaller in this experiment compared to previous experiments. Therefore the DA training is partially transferable from black images to MNIST.

\subsection{Characterization of the effect of domain adversarial against blur perturbation}

\begin{figure}
    \centering
    \includegraphics[scale = 0.3]{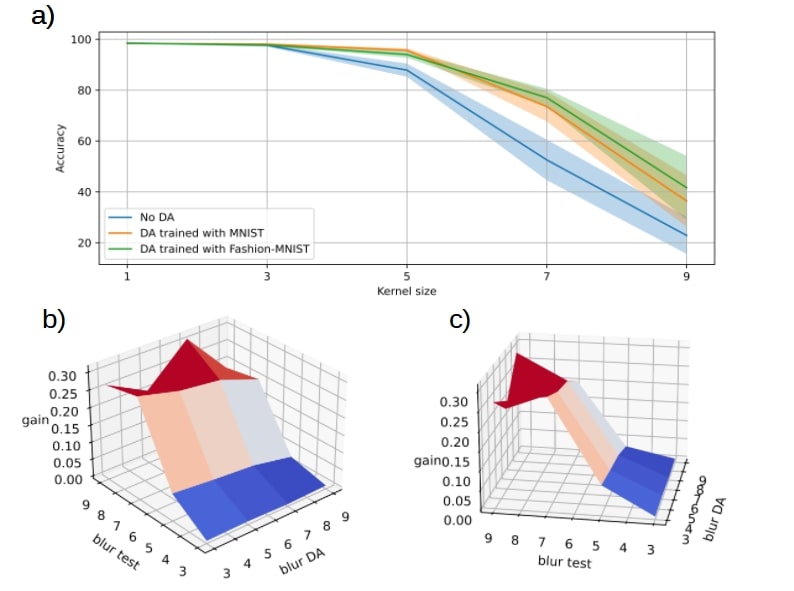}
    \caption{a) Representation of accuracy for image classification of MNIST dataset based on blur intensity. For each curve DA is trained with a different dataset. The blur intensity in DA is the same as in the test dataset. b) Representation of the gain with DA trained with Fashion-MNIST and blur as data shift. c) Same image as b) with different orientation.}
    \label{fig:MNIST_blur}
\end{figure}

In a second step we replaced noise by blur as data shift. Blur is a perturbation that often occurs in digitized histopathological samples. Here, we use MNIST for the train and test datasets, and also MNIST for the DA. The reference accuracy is 0.98 and there is no DA cost. There is a sigmoidal relation between blur intensity and performance degradation and accuracy falls to a minimum accuracy of 0.23 (degradation of 0.75) for a kernel size of 9. Once more, the gain is mainly sensitive to the blur intensity as a Gaussian function. The maximum gain is 0.25 for a kernel size of 7 in the dataset test and of 3 for the DA (Figure \ref{fig:MNIST_blur}).
\\
\\
We then replaced the MNIST dataset by a Fashion-MNIST dataset as DA datasets. We still have a degradation that can be modeled by a sigmoid function, we find a constant cost equal to zero and an important gain of 0.26. With similar results using Fashion-MNIST in the DA rather than MNIST, we showed that the inter-semantical DA transferability is applicable also with another data shift than noise, here blur.

\subsection{Characterization of the effect of domain adversarial training on histopathological datasets against blur perturbation and color shift}

\begin{figure}
    \centering
    \includegraphics[scale = 0.35]{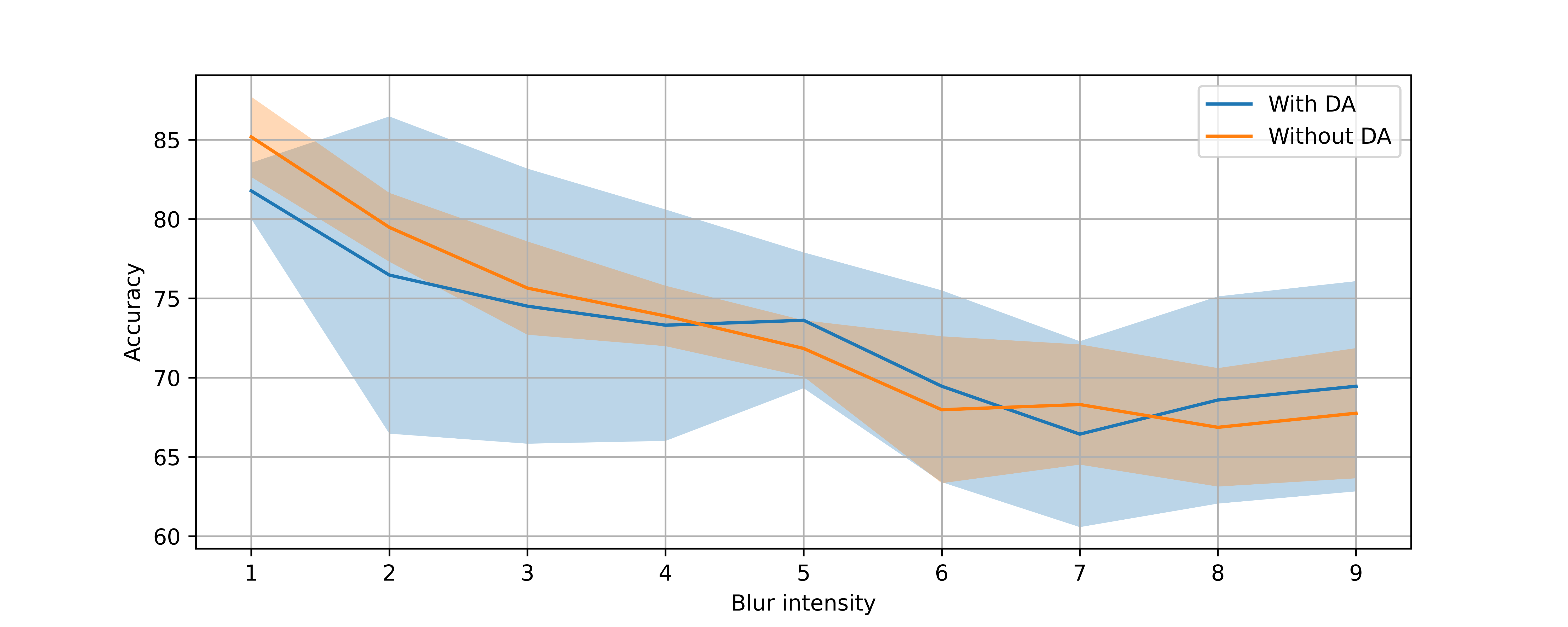}
    \caption{Comparison of the accuracy with and without DA training according to blur intensity as data shift on CAMELYON dataset. DA datasets are also CAMELYON datasets.}
    \label{fig:CAM_blur}
\end{figure}

We next used the CAMELYON dataset for both the main task and the DA and used blur as data shift as it is often found as natural perturbation in histopathological images. The reference accuracy is 0.85 and DA has a significant cost of 0.05 while there was no significant gain. This shows that DA can be deleterious depending on the processed dataset, and that the gain of DA is not ubiquitous. A possible explanation could be that blur erases important information from the histopathological data, therefore the model cannot recover from this kind of data degradation. Maximum blur with kernel size of 9x9 showed a degradation of 0.18 making accuracy fall to 0.67 (Figure \ref{fig:CAM_blur}).
\\
\\

We next studied color shift, another common shift from histopathological images, as data shift with the CAMELYON dataset. As color shift depends on more than one parameter, it is difficult to introduce a consistent intensity for a color shift. Thus, we create datasets of different domains of color shift that show different degradation by normalizing the dataset according to different reference images (Figure \ref{fig:dataset1}).
\\
\\
The reference has an accuracy of 0.85. The degradation varies from 0.03 to 0.26 depending on the domain shift. The cost varies from 0.0 to 0.10 with a small correlation with the amplitude of the degradation. As the cost is not null, it is important to carefully design the DA architecture and training process in order to prevent a loss of accuracy for the main task. The gain varies from 0.04 to 0.28, the gain is well correlated with the degradation. This is intuitive because the more performance degrades, the more DA training can be helpful.

\subsection{Characterization and inter-semantic transferability of the domain adversarial training on histopathological datasets against color shift}

\begin{figure*}
    \centering
    \includegraphics[scale = 0.4]{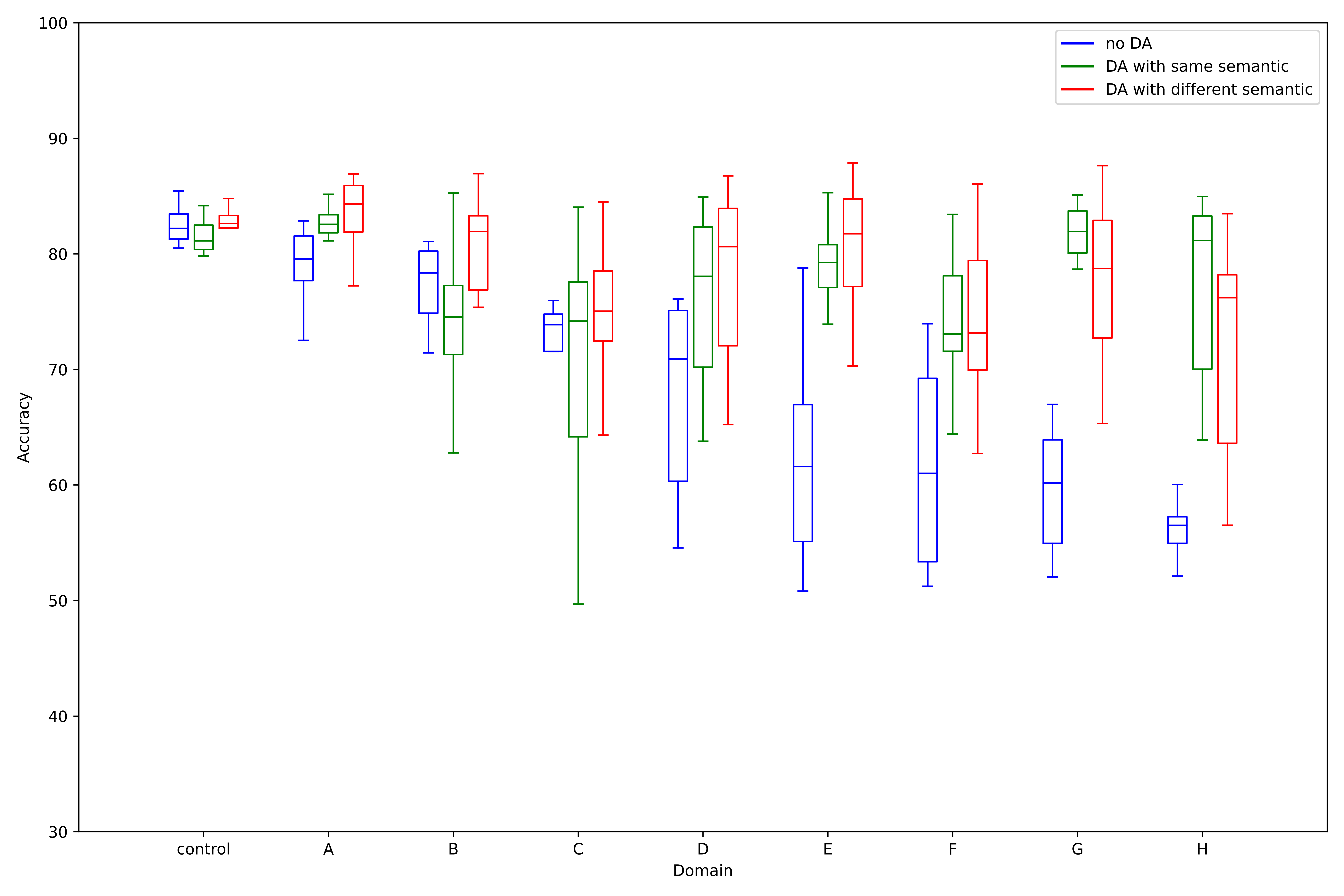}
    \caption{Box plot of the accuracy in different domains, the blue boxes are the accuracy without using the DA, the green boxes are the accuracy using a DA with the same semantics and the red boxes are the accuracy using a DA with different semantics. The datasets used for the DA training are normalized in the same domains as the test datasets.}
    \label{fig:CAM_stain}
\end{figure*}

We next used the TissueNet dataset for the DA training instead of the CAMELYON dataset, while keeping CAMELYON dataset for the main classification task and using color shift as data shift. This configuration is similar to the previous experiment using MNIST for the main task and Fashion-MNIST for the DA (Figure \ref{fig:CAM_stain}). However, there is no notion of data shift intensity for color shift because color shift is multidimensional.
\\
\\
In this configuration, the reference accuracy is 0.82. The degradation varies from 0.03 to 0.26 depending on the domain shift. The cost varies from 0.0 to 0.16 with a small correlation with the amplitude of the degradation. The gain varies from 0.02 to 0.28, the gain is well correlated with the degradation. Values of cost, degradation and gain are very close to the previous intra-semantic configuration, showing that DA training can be efficiently transferred inter-semantically in real histopathological images. Further investigations will be needed in order to understand if DA training can be applied to histopathological data with a gain greater than a cost universally, therefore increasing robustness of the model.

\section{Discussion and Conclusion}
Histopathological data is highly heterogeneous due to the diversity of acquisition devices and the lack of standard, while histopathological data is hardly available because many regulatory requirements are necessary to get access to clinical data. Together, lack of availability and heterogeneity are a major barrier for the development of safe and robust models. Therefore, we developed here a strategy to increase robustness using all available data diversity using domain adversarial methods.
\\
\\
By systematic analysis of DA effect over the model performance, we found that when DA is efficient, a low intensity in DA shift is sufficient to provide most of the possible gain from DA. But DA is not always efficient, blur degradation on histopathological datasets could not be retrieved using DA methods, this could be explained by two reasons: first, blur is already present in the original CAMELYON dataset therefore DA has no effect, and second, blur is a destructive noise which quickly makes classification impossible because relevant information may be found in high resolution patterns. Finally, inter-semantic DA transferability is an efficient strategy as it works using different dataset and with non-real data as the DA with black images showed an effective performance improvement.
\\
\\
In conclusion, DA training is transferable inter-semantically and the robustness of clinical algorithms can be increased by taking advantage of the heterogeneity of available datasets, whatever their semantic content is. Further investigation will be needed to understand when DA training is beneficial and when it is deleterious. Another remaining question is whether DA training can be done with inner inter-semantic datasets (data of different domains are also of different semantic). In this configuration, DA might erase features that are relevant for the main task. However, use of DA should be careful as it can significantly and negatively affect model performance.

%%%%%%
\newpage
\bibliographystyle{unsrt}
\bibliography{main}

\end{document}